\documentclass[sigconf]{acmart}
\usepackage{array}
\usepackage[ruled]{algorithm2e}
\usepackage{balance}
\usepackage{multirow}
\usepackage{subcaption}
\usepackage[normalem]{ulem}
\usepackage{graphicx}

\AtBeginDocument{%
  }

\begin{document}

\title[\textsc{Sera}: Semantic Representation Aggregation with Knowledge and LLM for Reliable and Interpretable Battery Health Forecasting]{\textsc{Sera}: Semantic Representation Aggregation with Knowledge and LLM for Reliable and Interpretable Battery Health Forecasting}

\makeatletter
\def\@affiliationfont{\fontsize{9.25pt}{11pt}\selectfont}  
\makeatother

\author{Jiawei Li}
\affiliation{
  \institution{Infocomm Technology Cluster}
  \institution{Singapore Institute of Technology}
  \country{Singapore}
  }
\email{jiawei.li@singaporetech.edu.sg} 

\author{Fang Liu}
\affiliation{
  \institution{School of Science and Technology}
  \institution{Singapore University of Social Sciences}
  \country{Singapore}}
\email{liufang@suss.edu.sg}

\author{Wei Zhang}
\authornote{Corresponding author.}
\affiliation{
  \institution{Infocomm Technology Cluster}
  \institution{Singapore Institute of Technology}
  \country{Singapore}}
\email{wei.zhang@singaporetech.edu.sg}

\author{Zuming Liu}
\affiliation{
  \institution{College of Smart Energy}
  \institution{Shanghai Jiao Tong University} 
  \country{Shanghai, China}}
\email{zmliu@sjtu.edu.cn}

\author{Man-Fai Ng}
\affiliation{
  \institution{Institute of Advanced Intelligence and Computing (IAIC), Agency for Science, Technology and Research (A*STAR), 1 Fusionopolis Way, \#16–16 Connexis, Singapore 138632, Republic of Singapore}
  \country{}}
\email{ngmf@a-star.edu.sg}

\author{Zhi Wei Seh}
\affiliation{
  \institution{Institute of Materials Research and Engineering (IMRE), Agency for Science, Technology and Research (A*STAR), 2 Fusionopolis Way, Innovis \#08-03, Singapore 138634, Republic of Singapore}
  \country{}}
\email{sehzw@a-star.edu.sg}


\renewcommand{\shortauthors}{J. Li et al.}

\begin{abstract}
Battery state of health (SoH) forecasting is important for battery management, but remains challenging due to nonlinear degradation and heterogeneity across batteries. Existing data-driven approaches primarily use temporal models to learn from numerical battery time series, and higher-level degradation characteristics are often not explicitly represented. These characteristics, however, can provide degradation guidance to support reliable forecasting and make the influence of degradation more interpretable. In this paper, we propose \textsc{Sera}, a \underline{se}mantic \underline{r}epresentation \underline{a}ggregation framework that complements temporal modelling with degradation semantics. Guided by battery domain expertise, \textsc{Sera} extracts degradation semantics from time series and constructs two complementary representations using rule-based knowledge and LLM-based interpretation. The representations are independently encoded and integrated with the representation learned by temporal models through gated aggregations. Experiments on the mainstream benchmark across multiple prediction horizons and different temporal models show that \textsc{Sera} consistently improves forecasting performance, achieving up to a 37.3\% reduction in prediction error over the temporal baseline and enhanced generalizability. Counterfactual analysis examines how forecasts respond to changes in degradation semantics to assess interpretability. The results show that prediction responses are consistent with the meanings of key degradation descriptors across tested horizons. Together, these findings demonstrate that structured degradation semantics and effective aggregation can improve forecasting accuracy and support reliable and interpretable battery health forecasting for advanced battery management.
\end{abstract}

\keywords{Battery health, semantic representation, large language model, representation learning, rule-based diagnostics.}

\maketitle

\section{Introduction}
\label{sec:introduction}
Predictive health management is increasingly important in manufacturing and industrial systems, where component health directly affects operational reliability, maintenance planning, and asset utilization. Batteries exemplify this need because their growing use in electric vehicles, energy storage, and industrial equipment requires monitoring and managing performance changes associated with repeated operation and electrochemical aging. Battery degradation reduces available capacity and operational efficiency, shortens service life, and, in severe cases, increases safety risks such as overheating and thermal runaway. These concerns make state of health (SoH), measured as the ratio of a battery’s current capacity to its initial capacity, an important indicator for characterizing battery health changes \cite{xiong2018towards,yang2024critical}. While SoH estimation characterizes the current battery condition, SoH forecasting predicts its future evolution to inform battery management and lifecycle planning \cite{semeraro2022battery}.

SoH forecasting is challenging because battery degradation is nonlinear, heterogeneous, and affected by multiple interacting factors. Electrochemical aging mechanisms, operating conditions, temperature, and usage patterns can correspond to different degradation trajectories \cite{vetter2005ageing,kabir2017degradation}. 
The trajectories differ even among batteries of the same model, limiting generalization of SoH models from one set of cells to unseen cells \cite{chen2024transfer,da2026conformalized}. The challenge becomes more evident when the prediction horizon increases, since the model must infer future degradation from a fixed amount of observed history. Besides, practical forecasting requires not only accurate temporal modelling but also representations that characterize the underlying degradation state to support reliable predictions across batteries and interpretable forecasting behavior.

Existing studies predict battery health using physics based models, data-driven methods, and hybrid solutions \cite{oji2021data,kannan2026smaller}. Deep sequential models, such as LSTM and its variants, learn degradation patterns from historical measurements through temporal modelling \cite{qu2019neural}, while Transformer based approaches use attention mechanisms to capture both spatial and temporal dependencies \cite{gu2023novel,long2025improved}. Transfer learning has also been investigated to improve generalization under domain shift \cite{ye2021state,duan2025lithium}. However, data-driven models often do not explicitly represent higher-level degradation characteristics, such as trend and severity, and possible degradation mechanisms. This limits the ability of the models to explicitly leverage such characteristics for prediction. As such, other approaches incorporate battery domain knowledge into SoH models. For example, empirical observations suggest that SoH generally declines with aging, electrical engineering provides equivalent circuit models to describe battery behavior, and materials science explains mechanisms such as lithium inventory loss and electrode degradation \cite{tan2026tide, qiu2026interpretable, li2025operando}. Such knowledge can be encoded through equations, parameters, or constraints, providing additional guidance and a more comprehensive understanding of battery degradation for SoH prediction.

Although temporal and knowledge-enhanced models have advanced battery health forecasting, existing methods predominantly represent degradation information numerically, which may not explicitly convey higher-level degradation characteristics. Our key idea is to introduce text-based semantic representations as another modality for characterizing degradation and enable more explicit and descriptive expression of degradation knowledge. These representations capture degradation trends and change points, severity and risk, and possible underlying mechanisms. By integrating semantic descriptions with numerical time series, we seek to leverage the strengths of different representations to improve forecasting accuracy and reliability and to enhance interpretability.

Based on this idea, we propose \textsc{Sera}, a \underline{se}mantic \underline{r}epresentation \underline{a}ggregation framework for reliable and interpretable battery health forecasting with both temporal and semantic representations. Given the battery time series data and domain knowledge, \textsc{Sera} constructs two semantic representations. The rule-based representation provides consistent degradation interpretation through predefined rules, and the LLM-based representation interprets relationships among multiple rules. These representations are encoded separately from the temporal representation and integrated through gated aggregations, which adaptively control the contribution of each representation. The aggregated representation is used to predict future SoH across multiple prediction horizons. Experiments on a mainstream benchmark show that \textsc{Sera} improves forecasting accuracy with up to a 37.3\% reduction in mean absolute error (MAE) over the temporal baseline and improved generalization to unseen batteries. Counterfactual analysis further shows that prediction responses are consistent with the meanings of key degradation descriptors, supporting the interpretability of the forecasting behavior.

The main contributions of this work are threefold. First, we introduce a structured semantic representation that translates battery degradation evidence into interpretable knowledge-derived descriptors and makes such information available to SoH models. Second, we develop \textsc{Sera}, which integrates rule- and LLM-based semantic representations with temporal modelling through gated aggregation, providing a general framework for adaptively incorporating semantic and temporal representations. Third, we evaluate \textsc{Sera} on a real-world dataset across multiple prediction horizons and temporal encoders, and use counterfactual semantic interventions to assess the consistency between forecast responses and the meanings of degradation semantics.

The remainder of this paper is organized as follows. Section \ref{sec:problem} defines the forecasting problem. Section \ref{sec:method} presents \textsc{Sera}. Section \ref{sec:experiment} reports the experimental setup, forecasting results, and interpretability analyses. Section \ref{sec:conclusion} concludes the paper.

\section{Battery SoH Forecasting Problem Formulation}
\label{sec:problem}
In this section, we formulate battery health forecasting as a multivariate sequence-to-sequence prediction problem and investigate the forecasting task for a future horizon. For a battery with $T$ observed cycles, each comprising charging and discharging time series, let $\mathbf{x}_t \in \mathbb{R}^{F}$ denote the feature vector extracted from cycle $t$, where $F$ is the number of input features. These features include cycle-level battery measurements and derived health indicators, such as voltage, current, temperature, and internal resistance. 
Given the most recent $L$ cycles ending at cycle $t \geq L$, the historical observation window is defined as,
\begin{equation}
\mathbf{X}_{t-L+1:t} = [\mathbf{x}_{t-L+1}, \mathbf{x}_{t-L+2}, \ldots, \mathbf{x}_{t}] \in \mathbb{R}^{L \times F},
\end{equation}
where $L$ denotes the observation window length. Let $y_t$ denote the SoH of the battery at cycle $t$. The forecasting target is the future SoH trajectory over the next $H$ cycles,
\begin{equation}
\mathbf{y}_{t+1:t+H} = [y_{t+1}, y_{t+2}, \ldots, y_{t+H}] \in \mathbb{R}^{H}.
\end{equation}
The forecasting model therefore learns the mapping,
\begin{equation}
\hat{\mathbf{y}}_{t+1:t+H} = f_{\theta}\left(\mathbf{X}_{t-L+1:t}\right),
\end{equation}
where $f_{\theta}(\cdot)$ denotes a forecasting model parameterized by $\theta$. The model predicts the future SoH trajectory as $\hat{\mathbf{y}}_{t+1:t+H}$, which is expected to closely approximate the ground truth trajectory $\mathbf{y}_{t+1:t+H}$.

\section{\textsc{Sera} Methodology} 
\label{sec:method}
In this section, we introduce the overall architecture of \textsc{Sera}, followed by semantic representation construction, temporal modelling, aggregation, and model training.

\subsection{Design Overview of \textsc{Sera}}
\begin{figure}[t]
\centering
\includegraphics[width=\columnwidth]{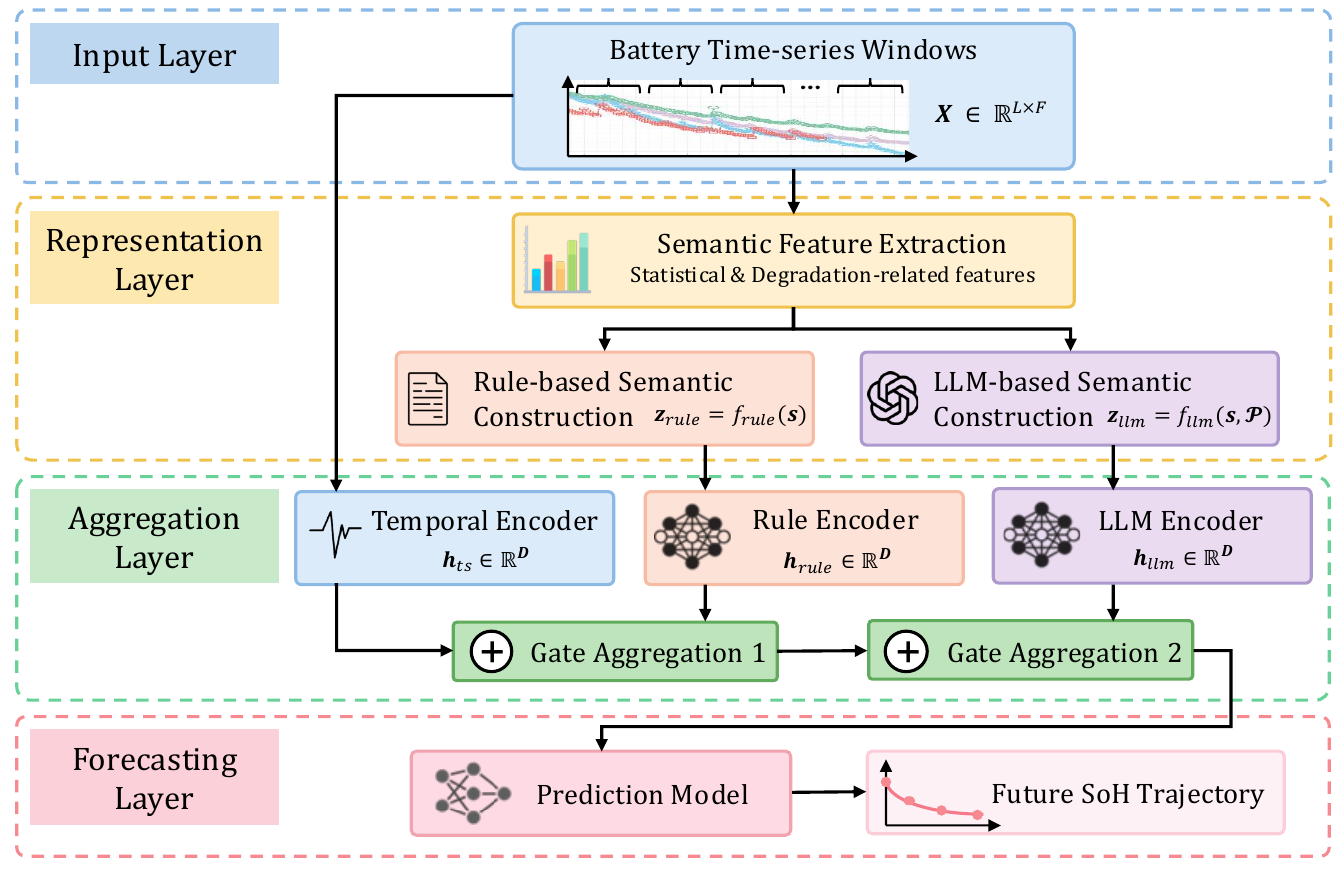}
\caption{Overall architecture of \textsc{Sera}. Given a window of battery time series, semantic features are extracted to construct rule-based and LLM-based degradation semantics. Temporal and semantic information is encoded separately and then aggregated for SoH forecasting.}
\label{fig:framework}
\end{figure}

The overall architecture of \textsc{Sera} is illustrated in Fig. \ref{fig:framework}, following a four-layer pipeline comprising input, representation, aggregation, and forecasting. At the input layer, a window of time series provides battery measurements that are processed through three branches. The temporal branch learns degradation patterns directly from numerical measurements and provides the baseline for forecasting. The rule-based branch applies battery domain knowledge to extracted degradation evidence to construct semantic descriptions of the battery state. The LLM-based branch interprets the degradation evidence to generate contextual descriptions within the same semantic schema. The representations from the three branches are then adaptively aggregated to predict future SoH.

Semantic representation is one of the key contributions of \textsc{Sera} and is constructed through the rule- and LLM-based branches at the representation layer. The rule-based branch constructs deterministic degradation descriptions using predefined rules grounded in battery domain knowledge. These rules are expected to provide consistent and useful degradation information to guide SoH forecasting. However, they are fixed and not exhaustive, and therefore offer limited flexibility in capturing diverse battery dynamics. To address this limitation, the LLM-based branch leverages the language understanding capability of LLMs to interpret degradation evidence more flexibly and provide richer semantic representations.

At the aggregation layer, the temporal, rule-based, and LLM-based information is independently encoded and integrated through a gated aggregation. The aggregation design enables \textsc{Sera} to adaptively control the contribution of different information sources rather than combining them deterministically. Finally, the aggregated representation is passed to the forecasting layer, where the prediction module generates the future SoH trajectory within the specified prediction horizon. The components of \textsc{Sera} are described in detail below.

\subsection{Structured Semantic Representation}
\label{sec:semantic_representation}

\begin{table*}[t]
\centering
\caption{Structured semantic representation schema for battery degradation. Each semantic component is described by its candidate values, supporting source features, and degradation meaning.}
\label{tab:semantic_schema}
\footnotesize
\renewcommand{\arraystretch}{1.1}
\begin{tabular}{p{0.10\textwidth}
                >{\raggedright\arraybackslash}p{0.18\textwidth}
                p{0.20\textwidth}
                p{0.35\textwidth}}
\hline
\textbf{Rule} &
\textbf{Candidate Values} &
\textbf{Source Features} &
\textbf{Description} \\
\hline

Trend &
\texttt{stable}, \texttt{mild\_decay},
\texttt{linear\_decay},
\texttt{accelerating\_fade} &
SoH slope, SoH acceleration ratio &
Overall degradation tendency observed within the input window. \\
\hline

Risk Level &
\texttt{low}, \texttt{medium}, \texttt{high} &
SoH slope, temperature std. ($T_{\mathrm{std}}$),
IR std. ($\sigma_{\mathrm{IR}}$) &
Estimated level of degradation risk based on evidence from
multiple battery indicators. \\
\hline

Mechanism &
\texttt{SEI\_growth}, \texttt{lithium\_plating},
\texttt{cathode\_degradation}, \texttt{thermal\_aging},
\texttt{impedance\_rise}, \texttt{mixed}, \texttt{stable} &
SoH slope, $\sigma_{\mathrm{IR}}$, $T_{\mathrm{std}}$,
charge time slope &
Dominant electrochemical or thermal degradation behaviour
inferred from the observed feature patterns. \\
\hline

Severity &
\texttt{mild}, \texttt{moderate}, \texttt{severe} &
Magnitude of SoH slope, mean SoH level &
Strength of the degradation evidence observed within the
input window. \\
\hline

Change Point &
\texttt{yes}, \texttt{no} &
SoH acceleration ratio, SoH acceleration flag &
Indication of a possible transition in the observed
degradation behaviour. \\
\hline

Key Metric &
\texttt{SoH\_slope}, \texttt{Tstd}, \texttt{IRstd},
\texttt{CTslp}, \texttt{QDslp} &
Selected diagnostic indicator &
Most informative feature supporting the current degradation
interpretation. \\
\hline

Metric Value &
Numerical ($\mathbb{R}$) &
Corresponding raw feature value &
Numerical value of the selected diagnostic metric obtained
from the extracted battery features. \\
\hline

\end{tabular}
\end{table*}

Guided by battery domain knowledge, we select seven semantic rules to characterize battery degradation, including trend, risk level, mechanism, severity, change point, key metric, and metric value. The rule definitions, candidate values, and associated degradation features are summarized in Table \ref{tab:semantic_schema}. The rules for trend and change point characterize degradation progression and its changes, while those for risk level and severity indicate degradation risk and severity. The mechanism rule describes possible underlying degradation processes, while key metric and metric value associate the semantic descriptions with quantitative degradation features. Together, these rules define the structured semantic representation as,
\begin{equation}
\mathbf{z} =
[z_{\mathrm{trend}}, z_{\mathrm{risk}}, z_{\mathrm{mech}},
z_{\mathrm{sev}}, z_{\mathrm{cp}},
z_{\mathrm{metric}}, z_{\mathrm{value}}].
\label{eq:semantic_schema}
\end{equation}

To assign values to these semantic rules, we preprocess the battery time series into a feature vector $\mathbf{s}$ that captures degradation characteristics required for semantic construction. These characteristics are derived from temporal changes and statistical variations over the observation window. Specifically, we compute the slope and acceleration ratio of the capacity-based SoH sequence to characterize degradation progression, together with statistical summaries of temperature and internal resistance to characterize operating variability. These derived features provide the quantitative basis for assigning values to the semantic rules. For example, the SoH slope and acceleration ratio are used to distinguish among \texttt{stable}, \texttt{mild\_decay}, \texttt{linear\_decay}, and \texttt{accelerating\_fade} for the trend rule. The resulting $\mathbf{s}$ provides a common input to both semantic construction branches.

The rule-based branch implements the semantic rules using predefined thresholds grounded in domain knowledge. Specifically, the preprocessed features in $\mathbf{s}$ are compared against their corresponding thresholds to determine the candidate values described above. This threshold-based assignment produces a deterministic semantic representation for a given feature vector. The mapping is expressed as,
\begin{equation}
\mathbf{z}^{\mathrm{rule}} = f_{\mathrm{rule}}(\mathbf{s}),
\label{eq:rule_semantics}
\end{equation}
where $f_{\mathrm{rule}}(\cdot)$ applies the same predefined rules across all input windows. Although this provides consistent semantic guidance, fixed thresholds may not capture contextual relationships among features, motivating the LLM-based branch.

The LLM branch interprets the same feature vector $\mathbf{s}$ through a prompt specification $\mathcal{P}$ that defines the semantic rules and their candidate values. The prompt instructs the LLM to consider the extracted features jointly when assigning values to the semantic rules, while restricting its output to the common schema, yielding,
\begin{equation}
\mathbf{z}^{\mathrm{llm}} =
f_{\mathrm{llm}}(\mathbf{s},\mathcal{P}),
\label{eq:llm_semantics}
\end{equation}
where $f_{\mathrm{llm}}(\cdot)$ denotes the LLM-based semantic mapping process. Drawing on knowledge learned from natural language, this branch provides greater flexibility in capturing contextual relationships among degradation features than predefined threshold-based mappings in Eq. \eqref{eq:rule_semantics}. The resulting representation provides complementary semantic guidance while retaining the same semantic rules and candidate values as the rule-based branch.

\subsection{Representation Encoding and Aggregation}
\label{sec:encoding_aggregation}
Given the raw battery time series and the rule- and LLM-based semantic representations, \textsc{Sera} encodes each input along its corresponding branch to obtain temporal and semantic embeddings. These three embeddings are then integrated through gated aggregation for subsequent SoH prediction.

\subsubsection{Temporal Encoding}
\label{sec:time_series_representation}
In the first branch, we perform temporal modelling using an encoder that processes the battery time series window $\mathbf{X}\in\mathbb{R}^{L\times F}$ to obtain a temporal embedding. We use PatchTST \cite{nie2022time} as the temporal encoder in this study. Its patch-based design groups consecutive observations into local segments, preserving patterns within each patch while reducing the number of tokens processed by attention compared with using one token per cycle. Specifically, the input window is divided into $M$ patches of length $P$, each projected into a $D$-dimensional embedding. A learnable classification (CLS) token, used here to summarize the input sequence, is prepended to the patch embeddings, and positional information is added to retain their temporal order. The resulting sequence of size $(M+1)\times D$ is processed by $L_{\mathrm{enc}}$ Transformer encoder layers. The final CLS token output provides the temporal embedding as,
\begin{equation}
\mathbf{h}_{\mathrm{ts}} =
\mathbf{E}_{\mathrm{cls}}^{(L_{\mathrm{enc}})}
\in \mathbb{R}^{D}.
\label{eq:temporal_representation}
\end{equation}
Note that \textsc{Sera} is not restricted to PatchTST and can accommodate other temporal encoders to generate temporal embeddings. We evaluate several alternative architectures in the experiments.



\subsubsection{Semantic Representation Encoding}
\label{sec:semantic_encoding}
The second and third branches independently encode the rule- and LLM-based semantic representations. These representations are generated through different methods and provide different interpretations of the degradation evidence. We use two separate encoders with the same architecture and independent parameters to retain their distinct information. In each branch, the structured semantic text is first encoded into a numerical vector, then processed by an MLP projection, a ReLU activation, and a final linear mapping to obtain the semantic embedding. The resulting semantic embeddings are,
\begin{equation}
\mathbf{h}_{\mathrm{rule}}=\texttt{Linear}\left(\texttt{ReLU}\left(\texttt{Proj}\left(\texttt{Embed}\left(\mathbf{z}^{\mathrm{rule}}\right)\right)\right)\right)\in \mathbb{R}^{D},
\label{eq:rule_encoder}
\end{equation}
\begin{equation}
\mathbf{h}_{\mathrm{llm}}=\texttt{Linear}\left(\texttt{ReLU}\left(\texttt{Proj}\left(\texttt{Embed}\left(\mathbf{z}^{\mathrm{llm}}\right)\right)\right)\right)\in \mathbb{R}^{D}.
\label{eq:llm_encoder}
\end{equation}
Here, $\texttt{Embed}(\cdot)$ converts the structured semantic representation into a numerical vector, and $\texttt{Proj}(\cdot)$ denotes the MLP projection with intermediate dimension $d_{\mathrm{proj}}=64$. The final linear mapping produces semantic embeddings of dimension $D$, matching the temporal embedding $\mathbf{h}_{\mathrm{ts}}$ for subsequent gated aggregation.

\subsubsection{Adaptive Gated Aggregation}
\label{sec:gated_aggregation}
\textsc{Sera} integrates the temporal and semantic embeddings through adaptive gated aggregation. The three embeddings are adaptively weighted and aggregated as,
\begin{equation}
\mathbf{h}_{\mathrm{agg}}
=
\mathbf{g}_{\mathrm{ts}}\odot\mathbf{h}_{\mathrm{ts}}
+\mathbf{g}_{\mathrm{rule}}\odot\mathbf{h}_{\mathrm{rule}}
+\mathbf{g}_{\mathrm{llm}}\odot\mathbf{h}_{\mathrm{llm}},
\label{eq:gated_aggregation}
\end{equation}
where $\mathbf{g}_{k}\in\mathbb{R}^{D}$ is the input-dependent gating vector for embedding $\mathbf{h}_{k}$, with $k\in\{\mathrm{ts},\mathrm{rule},\mathrm{llm}\}$. Each gating vector is computed as $\mathbf{g}_{k}=\sigma(\mathbf{W}_{k}\mathbf{h}_{k})$, where $\mathbf{W}_{k}\in\mathbb{R}^{D\times D}$ contains learned parameters and $\sigma(\cdot)$ denotes the \texttt{sigmoid} function. The operator $\odot$ denotes element-wise multiplication. Each gate assigns a factor between zero and one to each embedding dimension, allowing the contributions of the three representations to adapt to individual input windows. Factors approaching zero suppress the corresponding dimensions, and factors approaching one retain them. The aggregated embedding $\mathbf{h}_{\mathrm{agg}}$ is subsequently used for SoH prediction.

\subsection{SoH Forecasting and Model Training}
\label{sec:forecast_training}
The aggregated embedding $\mathbf{h}_{\mathrm{agg}}$ is passed to an MLP prediction head to generate the future SoH trajectory over $H$ cycles as,
\begin{equation}
\hat{\mathbf{y}}_{t+1:t+H}
=\mathbf{W}_{\mathrm{out}}\texttt{ReLU}\left(
\mathbf{W}_{\mathrm{hid}}\mathbf{h}_{\mathrm{agg}}
+\mathbf{b}_{\mathrm{hid}}\right)+\mathbf{b}_{\mathrm{out}},
\label{eq:prediction_head}
\end{equation}
where $\mathbf{W}_{\mathrm{hid}}$, $\mathbf{W}_{\mathrm{out}}$ and their corresponding biases are learnable parameters, and $\hat{\mathbf{y}}_{t+1:t+H}\in\mathbb{R}^{H}$. The model is trained using a composite objective that combines prediction accuracy with constraints on the temporal behaviour of the predicted trajectory,
\begin{equation}
\mathcal{L}=\mathcal{L}_{\mathrm{base}}
+\lambda_{\mathrm{mono}}\mathcal{L}_{\mathrm{mono}}
+\lambda_{\mathrm{smooth}}\mathcal{L}_{\mathrm{smooth}},
\label{eq:total_loss}
\end{equation}
where $\lambda_{\mathrm{mono}}$ and $\lambda_{\mathrm{smooth}}$ control the contributions of the two penalty terms. The base term uses a horizon-weighted \texttt{SmoothL1} (Huber) loss that assigns larger weight to predictions farther into the future, reflecting the increasing difficulty of longer-horizon forecasting.
\begin{equation}
\mathcal{L}_{\mathrm{base}}
=\frac{1}{BH}\sum_{i=1}^{B}\sum_{j=1}^{H}
w_j\,\texttt{SmoothL1}_{\beta}(\hat{y}_{i,j},y_{i,j}),
\ w_j=\left(\frac{j}{H}\right)^{\alpha},
\label{eq:base_loss}
\end{equation}
where $B$ is the batch size, $\beta$ is the transition threshold of the \texttt{SmoothL1} loss, and $\alpha>0$ controls the weighting across the prediction horizon. Larger weights are assigned to later prediction steps, which encourages the model to maintain forecasting accuracy as the prediction distance increases. Since battery SoH generally declines with aging, a monotonicity penalty discourages increases between consecutive predictions, and we have,
\begin{equation}
\mathcal{L}_{\mathrm{mono}}
=\frac{1}{B(H-1)}\sum_{i=1}^{B}\sum_{j=1}^{H-1}
\max\left(0,\hat{y}_{i,j+1}-\hat{y}_{i,j}\right).
\label{eq:mono_loss}
\end{equation}
Electrochemical aging accumulates slowly over multiple cycles, so that battery SoH typically changes gradually. A smoothness penalty discourages large changes between consecutive predictions with,
\begin{equation}
\mathcal{L}_{\mathrm{smooth}}
=\frac{1}{B(H-1)}\sum_{i=1}^{B}\sum_{j=1}^{H-1}
\left|\hat{y}_{i,j+1}-\hat{y}_{i,j}\right|.
\label{eq:smooth_loss}
\end{equation}
Together, these loss terms encourage accurate forecasts with monotonically and gradually decreasing SoH trajectories.

\section{Experiments}
\label{sec:experiment}
This section evaluates the forecasting performance of \textsc{Sera}. We first describe the experimental setup, followed by the main results and further analyses.
    
\subsection{Experimental Setup}
\label{sec:experimental_setup}
We describe the dataset and sample construction, implementation settings of \textsc{Sera}, and comparison variants.

\subsubsection{Dataset and Forecasting Sample Construction}
\label{sec:datasets}
We evaluate \textsc{Sera} on the MIT Battery Degradation Dataset \cite{severson2019data}, which contains 124 commercial LFP/graphite cells cycled to end-of-life under fast charging conditions at $30^\circ$C. The tested cells are A123 Systems APR18650M1A cells with a nominal capacity of 1.1 Ah and a nominal voltage of 3.3 V. The dataset provides cycle-level measurements of capacity, voltage, current, temperature, and internal resistance throughout the battery degradation process. 

The batteries are divided by cell identity into 70\%, 15\%, and 15\% for training, validation, and testing, respectively, ensuring that no cell appears in more than one set.
Forecasting samples are generated from each battery trajectory using sliding windows. For a battery trajectory of length $T$, a valid sample requires both an $L$-cycle observation window and its subsequent $H$-cycle forecasting target. The $i$-th sample is constructed as,
\begin{equation}
\mathcal{W}_i = \left( \mathbf{X}_{i:i+L-1}, \mathbf{y}_{i+L:i+L+H-1} \right),
\end{equation}
where $i=1+kw$ for nonnegative integers $k$ satisfies $i\leq T-L-H+1$, and $w$ is the sliding stride. We set $w=1$, advancing the window by one cycle and producing $N=T-L-H+1$ samples for each battery with $T\geq L+H$. For example, with $L=30$ and $H=90$, the first sample uses cycles 1--30 to predict cycles 31--120.

We fix the look-back length at $L=30$ cycles and evaluate prediction horizons $H\in\{30,50,70,90\}$, keeping the observed history unchanged. The relative forecasting distance is characterized by the forecast-to-observation ratio,
\begin{equation}
    \rho = \frac{H}{L}.
\end{equation}
The four forecasting settings correspond to $\rho \in \{1.0,1.67,2.33,3.0\}$. Increasing $\rho$ extends the forecast farther beyond the observation window, allowing us to assess forecasting robustness across prediction distances under a fixed history length.

\subsubsection{\textsc{Sera} Configurations}
\label{sec:sera_config}
For the temporal branch, PatchTST is used with patch length $P=5$ and representation dimension $D=128$. The Transformer module of PatchTST $L_{\mathrm{enc}}$ contains three encoder layers with four attention heads. For the semantic branches, the rule- and LLM-based representations use separate encoders with the same schema, where the intermediate projection dimension $d_{\mathrm{proj}}$ is set to 64. The resulting semantic embeddings are mapped to the same dimension $D$ before adaptive gated aggregation.
All models are trained using \texttt{Adam} with a learning rate of $10^{-3}$ and a batch size of 32. Training is performed for up to 80 epochs with gradient clipping at 1.0 and early stopping based on the validation loss with a patience of 10 epochs. For the training objective, the horizon weighting parameter is set to $\alpha=1.0$, while $\lambda_{\mathrm{mono}}=0.1$ and $\lambda_{\mathrm{smooth}}=0.05$. The \texttt{SmoothL1} threshold $\beta$ is set to 0.01. The same training configuration is used across all prediction horizons for consistent evaluation. 

\subsubsection{Comparison Models}
\label{sec:comparison_variants}
We compare \textsc{Sera} with several comparison models to evaluate the contribution of different semantic sources. TS uses only the time series or temporal representation and serves as the forecasting baseline. TS + Rule augments the temporal representation with rule-based semantics, while TS + LLM incorporates LLM-based semantics. \textsc{Sera} represents the complete model, which aggregates both rule- and LLM-based semantic representations with the temporal representation. We also include Rule only and LLM only to examine whether semantic representations can support forecasting without temporal information. All variants use the same input features, data partitions, and training settings for consistent comparison.

To examine whether the proposed semantic representation and aggregation approach depends on a particular temporal model, we further evaluate \textsc{Sera} with PatchTST, a standard Transformer (TFM) \cite{vaswani2017attention}, and LSTM as the temporal encoders. Forecasting performance is measured using MAE and root mean squared error (RMSE). A single NVIDIA RTX 4090 GPU is used.

\subsection{Forecasting Performance Analysis}
\label{sec:forecasting_performance}
We report the experimental results and evaluate the forecasting performance of \textsc{Sera} under the following settings.

\begin{table}[t]
\centering
\caption{Comparison of forecasting performance in terms of test MAE and RMSE across different prediction horizons. Lower values indicate better performance, with the best results highlighted in bold. All values are $\times 10^{-3}$.}
\label{tab:main_results}
\footnotesize
\renewcommand{\arraystretch}{1.08}
\setlength{\tabcolsep}{3.5pt}
\begin{tabular}{lcccccccc}
\toprule
& \multicolumn{2}{c}{$H=30$}
& \multicolumn{2}{c}{$H=50$}
& \multicolumn{2}{c}{$H=70$}
& \multicolumn{2}{c}{$H=90$} \\
\cmidrule(lr){2-3}
\cmidrule(lr){4-5}
\cmidrule(lr){6-7}
\cmidrule(lr){8-9}

Model
& MAE & RMSE
& MAE & RMSE
& MAE & RMSE
& MAE & RMSE \\
\midrule

TS
& 1.556 & 2.309
& 2.317 & 2.942
& 3.043 & 3.818
& 3.565 & 4.727 \\ \midrule

Rule only
& 11.384 & 12.904
& 12.064 & 13.578
& 10.962 & 12.174
& 11.641 & 13.172 \\

LLM only
& 12.157 & 14.225
& 12.806 & 14.755
& 11.526 & 13.296
& 12.150 & 14.515 \\ \midrule

TS + Rule
& \textbf{0.912} & 1.369
& 1.883 & 2.463
& 2.350 & 3.135
& 2.593 & 3.349 \\

TS + LLM
& 1.408 & 2.036
& 2.108 & 2.744
& 2.522 & 3.202
& 2.946 & 3.869 \\ \midrule

\textsc{Sera} (ours)
& 1.015 & \textbf{1.323}
& \textbf{1.688} & \textbf{2.122}
& \textbf{2.142} & \textbf{2.851}
& \textbf{2.234} & \textbf{2.971} \\

\bottomrule
\end{tabular}
\end{table}

\subsubsection{Overall Forecasting Performance}
\label{sec:overall_performance}
We first evaluate whether incorporating structured degradation semantics improves SoH forecasting beyond the temporal representation alone. Table \ref{tab:main_results} presents the results of \textsc{Sera}, the temporal baseline, and different semantic configurations across four prediction horizons, using PatchTST as the temporal encoder.
Overall, \textsc{Sera} achieves the lowest RMSE at all tested horizons and the lowest MAE for long horizons with $H\in\{50,70,90\}$. Compared with baseline TS, \textsc{Sera} reduces MAE from 1.6 to 1.0, 2.3 to 1.7, 3.0 to 2.1, and 3.6 to 2.2 (all $\times 10^{-3}$) at $H=30$, 50, 70, and 90, respectively. These results indicate that structured degradation semantics complement the temporal representation with explicit text-based degradation information, helping the model better capture the evolving battery state and produce more accurate SoH predictions. An exception occurs at $H=30$, where TS + Rule achieves the lowest MAE of $0.9\times10^{-3}$, compared with $1.0\times10^{-3}$ for \textsc{Sera}. One possible explanation is that the temporal and rule-based representations already capture the dominant degradation information needed for shorter prediction horizon, while the additional LLM representation affects the efficiency of the forecasting module. As the prediction horizon increases, the comprehensive information provided by both semantic sources becomes more useful, and \textsc{Sera} achieves the best MAE and RMSE. 

The results also clarify the role of semantic information. Both TS + Rule and TS + LLM generally improve upon TS, showing that each semantic source provides useful degradation guidance. In contrast, Rule only and LLM only produce MAE values from $0.011$ to $0.014$, substantially higher than the models that utilize temporal information. This implies that semantic descriptors alone do not retain sufficient information about the absolute SoH level and data-centric trajectory dynamics. They therefore serve more effectively as complementary guidance to the temporal representation than as standalone predictors.

\begin{figure}[t]
    \centering

    \begin{subfigure}[t]{0.49\columnwidth}
        \centering
        \includegraphics[width=\linewidth]
        {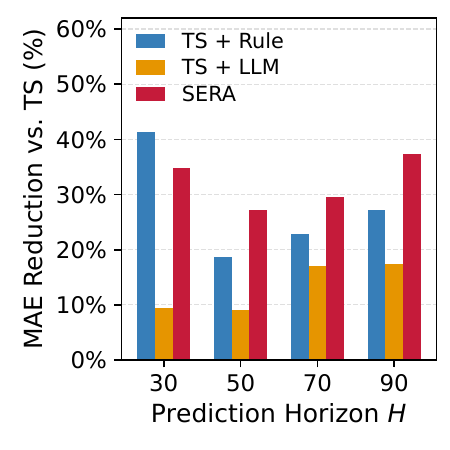}
        \caption{Semantic contribution}
        \label{fig:semantic_contribution}
    \end{subfigure}
    \hfill
    \begin{subfigure}[t]{0.49\columnwidth}
        \centering
        \includegraphics[width=\linewidth]
        {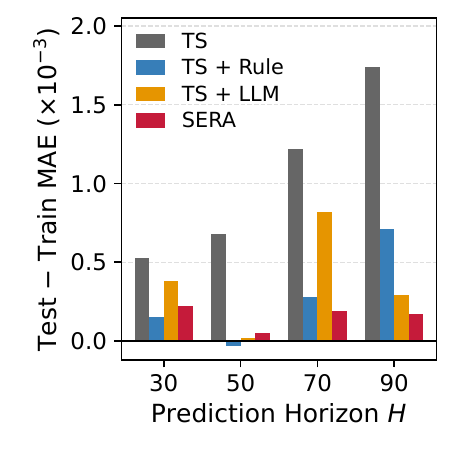}
        \caption{Generalization gap}
        \label{fig:generalization_gap}
    \end{subfigure}
    \caption{Semantic contribution and generalization analysis across prediction horizons. Fig. \ref{fig:semantic_contribution}: reduction in test MAE relative to the TS baseline. Fig. \ref{fig:generalization_gap}: generalization gap, measured as the difference between test and training MAE.}
\label{fig:semantic_analysis}
\end{figure}

\subsubsection{Semantic Contribution and Generalization}
\label{sec:semantic_generalization}
We further examine the contribution of semantic information to forecasting and the model’s generalization to unseen batteries. Fig. \ref{fig:semantic_contribution} reports the percentage reduction in test MAE relative to the TS baseline, calculated as $R=(\mathrm{MAE}_{\mathrm{TS}}-\mathrm{MAE}_{\mathrm{variant}})/
\mathrm{MAE}_{\mathrm{TS}}\times100\%$. A larger positive value indicates a greater relative improvement over TS. Aligned with the results in Table \ref{tab:main_results}, the contribution of semantic information varies with the prediction horizon. At the short horizon $H=30$, TS + Rule achieves the largest MAE reduction of 41.4\%, compared with 34.8\% for \textsc{Sera}, suggesting that rule-based semantics already provide effective guidance for short term forecasting. For the relative longer horizons $H\in\{50,70,90\}$, \textsc{Sera} consistently outperforms both individual semantic configurations, and its MAE reduction increases from 27.1\% to 29.6\% and then to 37.3\%. Its advantage over TS + Rule is 8.4\%, 6.8\%, and 10.1\% across these horizons, respectively. These results suggest that combining rule- and LLM-based semantics and temporal representation becomes increasingly beneficial as the prediction horizon extends.

Fig. \ref{fig:generalization_gap} provides a different view through the gap between test and training MAE. A gap near zero suggests generalized performance on training and unseen batteries and a positive gap may indicate overfitting. The TS baseline shows a larger gap as $H$ increases, reaching $1.7\times10^{-3}$ at $H=90$, whereas \textsc{Sera} maintains a relatively small gap across all horizons. A possible reason is that structured semantics capture degradation characteristics shared across batteries, reducing the model's dependence on battery specific temporal patterns. The results suggest that semantic information supports both forecasting accuracy and generalization to unseen batteries.

\begin{table}[t]
\centering
\caption{Comparison of test MAE across temporal encoders at $H=30$ and $H=90$. Lower values are better, and the best results are highlighted in bold. All values are $\times 10^{-3}$.}
\label{tab:cross_temporal}
\footnotesize
\renewcommand{\arraystretch}{1.08}

\begin{tabular}{lcccccc}
\toprule
\multirow{2}{*}{Model}
& \multicolumn{2}{c}{LSTM}
& \multicolumn{2}{c}{TFM}
& \multicolumn{2}{c}{PatchTST} \\
\cmidrule(lr){2-3}
\cmidrule(lr){4-5}
\cmidrule(lr){6-7}
& $H=30$ & $H=90$
& $H=30$ & $H=90$
& $H=30$ & $H=90$ \\
\midrule

TS
& 1.394 & 3.578
& 1.599 & 3.421
& 1.556 & 3.565 \\ \midrule

TS + Rule
& 1.505 & \textbf{3.097}
& 1.375 & 2.708
& \textbf{0.912} & 2.593 \\

TS + LLM
& 1.535 & 3.336
& 2.013 & 3.246
& 1.408 & 2.946 \\ \midrule

\textsc{Sera} (ours)
& \textbf{1.229} & 3.098
& \textbf{1.350} & \textbf{2.636}
& 1.015 & \textbf{2.234} \\

\bottomrule
\end{tabular}
\end{table}

\subsubsection{Temporal Encoder Sensitivity}
\label{sec:cross_temporal_encoder}
We further examine the impact of temporal encoder choice. Table \ref{tab:cross_temporal} compares the TS baselines and corresponding comparison models using LSTM, TFM, and PatchTST as the temporal encoder at $H=30$ and $H=90$.
Although incorporating semantic representations improves forecasting across all three temporal encoders, the gains vary with the temporal encoder and prediction horizon. With LSTM as the temporal encoder, \textsc{Sera} achieves the lowest MAE at $H=30$ and performs nearly identically to TS + Rule at $H=90$. With TFM, \textsc{Sera} achieves the lowest MAE at both horizons. A similar benefit is observed with PatchTST, where \textsc{Sera} reduces the MAE from $3.6\times10^{-3}$ for TS to $2.2\times10^{-3}$ at $H=90$. These results suggest that the benefit of semantic representations is not specific to a particular temporal encoder and that semantic information can be effectively aggregated with different temporal representations. Meanwhile, the overall forecasting performance also varies across temporal encoders, with \textsc{Sera} generally achieving better performance with PatchTST than with LSTM and TFM. The varying gains further indicate that the effectiveness of semantic information depends partly on the temporal representations with which it is aggregated.

\begin{figure}[t]
    \centering
    \begin{subfigure}[t]{0.495\columnwidth}
        \centering
        \includegraphics[width=\linewidth]
        {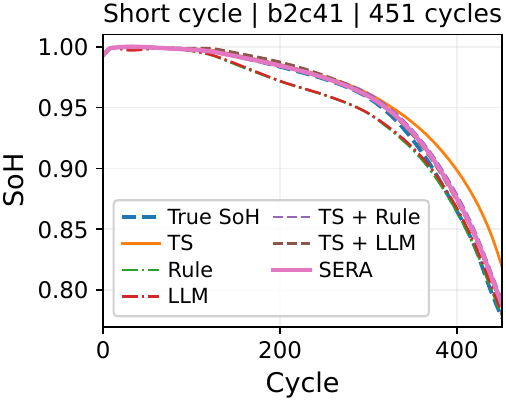}
        \caption{Short cycle battery}
        \label{fig:h90_short}
    \end{subfigure}
    \hfill
    \begin{subfigure}[t]{0.495\columnwidth}
        \centering
        \includegraphics[width=\linewidth]
        {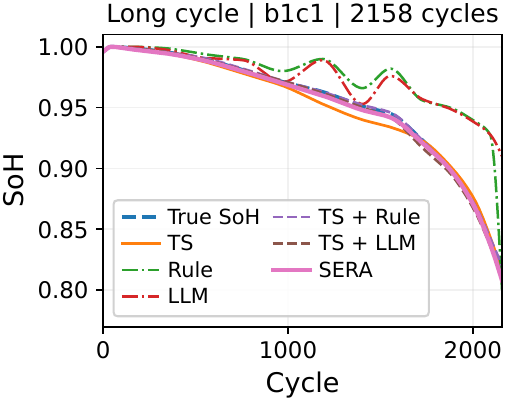}
        \caption{Long cycle battery}
        \label{fig:h90_long}
    \end{subfigure}
    \caption{Qualitative comparison of SoH trajectories for representative short- (Fig. \ref{fig:h90_short}) and long-lifespan (Fig. \ref{fig:h90_long}) batteries. The results compare the ground truth trajectories with predictions from the temporal baseline, semantic variants, and models with temporal and semantic representations.}    \label{fig:h90_case_study}
\end{figure}

\subsubsection{Case Study}
\label{sec:qualitative}
To complement the aggregate forecasting results, Fig. \ref{fig:h90_case_study} illustrates the full SoH trajectories of two representative batteries with short and long lifespans. The trajectories provide a detailed view of prediction behavior across degradation stages that is not reflected by aggregate error metrics. In both cases, \textsc{Sera} remains close to the ground truth SoH during gradual degradation and accelerated decline stages.
Specifically, for the short-lifespan case in Fig. \ref{fig:h90_short}, the trajectories predicted by different SoH models remain close during the early cycles. However, after approximately 300 cycles, TS gradually deviates from the ground truth and increasingly overestimates SoH, while \textsc{Sera} follows the accelerated decline more closely. For the long-lifespan battery in Fig. \ref{fig:h90_long}, the main difference occurs between approximately 800 and 1,600 cycles, where TS underestimates SoH while \textsc{Sera} better captures the gradual decline. These patterns illustrate how incorporating degradation semantics can help the model capture different degradation behaviors across battery life stages.
As also shown in the figures, the Rules and LLM curves fall below the ground truth during the middle stage in Fig. \ref{fig:h90_short}, while exhibiting larger fluctuations and a later transition to accelerated degradation in Fig. \ref{fig:h90_long}. These deviations suggest that semantic descriptors alone do not fully capture the numerical SoH trajectory, supporting their use to guide rather than replace the temporal representation in \textsc{Sera}.

\subsection{Interpretability Analysis}
\label{sec:explainability}
Beyond forecasting accuracy, it is important to interpret the influence of structured semantic representations on the predictions of \textsc{Sera} and we perform counterfactual semantic analysis.

\subsubsection{Counterfactual Semantic Analysis}
\label{sec:counterfactual_setup}
Counterfactual analysis examines whether changes in the semantic representation lead to expected changes in the forecast while the observed time series remains unchanged. This allows us to isolate the influence of semantic information from that of the temporal input and assess whether \textsc{Sera} uses the semantic descriptors consistently with their degradation meaning.
Given an input window $\mathbf{X}$ and its rule- and LLM-based semantic representations $\mathbf{z}^{\mathrm{rule}}$ and $\mathbf{z}^{\mathrm{llm}}$, the original prediction is,
\begin{equation}
\hat{\mathbf{y}}=f(\mathbf{X},\mathbf{z}^{\mathrm{rule}},\mathbf{z}^{\mathrm{llm}}).
\label{eq:cf_original}
\end{equation}
We then modify one or multiple rules in both semantic representations while keeping $\mathbf{X}$ fixed, producing counterfactual representations $\mathbf{z}^{\mathrm{rule}'}$ and $\mathbf{z}^{\mathrm{llm}'}$. Applying the same intervention to both representations maintains a consistent hypothetical degradation state. The intervention changes the corresponding prediction to,
\begin{equation}
\hat{\mathbf{y}}'=f(\mathbf{X},\mathbf{z}^{\mathrm{rule}'},\mathbf{z}^{\mathrm{llm}'}).
\label{eq:cf_prediction}
\end{equation}
Here, the intervention effect at prediction step $t'$ is measured by,
\begin{equation}
\Delta_{t'}=\hat{y}'_{t+t'}-\hat{y}_{t+t'},\quad t' \in \{1,\ldots,H\}.
\label{eq:cf_shift}
\end{equation}
A positive value indicates an increase in predicted SoH after the intervention, and a negative value indicates a decrease. For example, changing the value of the Trend rule from \texttt{stable} in $\mathbf{z}^{\mathrm{rule}}$ and $\mathbf{z}^{\mathrm{llm}}$ to \texttt{accelerating\_fade} in $\mathbf{z}^{\mathrm{rule}'}$ and $\mathbf{z}^{\mathrm{llm}'}$ is expected to lower predicted SoH, as battery degradation typically accelerates near the battery's end of life with low SoH. If the predicted $\hat{\mathbf{y}}'$ with such intervention decreases relative to the predicted $\hat{\mathbf{y}}$ without intervention, the response agrees with the expectation based on battery domain knowledge. We call this agreement semantic consistency and use it to interpret the alignment of the model’s response with the rule’s degradation meaning.

We evaluate 11 interventions covering individual semantic fields, degradation slope, and coherent optimistic or pessimistic combinations. For each prediction horizon, each intervention is evaluated over $n_{\mathrm{cf}}=200$ randomly sampled counterfactual test windows. We report semantic consistency as the percentage of windows whose predictions change in the expected direction. We also report the mean prediction shift at the final cycle of the horizon as,
\begin{equation}
\bar{\Delta}_{H} = \frac{1}{n_{\mathrm{cf}}} \sum_{i=1}^{n_{\mathrm{cf}}}\Delta_{H}^{i},
\label{eq:cf_mean_shift}
\end{equation}
where $\Delta_{H}^{i}$ is the SoH shift for the $i$-th sampled window. The sign of $\bar{\Delta}_{H}$ shows the average SoH response direction, while $|\bar{\Delta}_{H}|$ measures the magnitude of the intervention effect.

\subsubsection{Semantic Consistency}
\label{sec:semantic_consistency}
Table \ref{tab:counterfactual} reports the semantic consistency and final-cycle shift in predicted SoH for each intervention. We first consider the optimistic bundle intervention, which jointly modifies multiple semantic rules toward a healthier degradation state, including a stable trend, low risk, a stable mechanism, and a mild SoH decline. Conversely, the pessimistic bundle intervention modifies these rules toward a worsening degradation state, including an accelerating trend, high risk, Li plating, and a steep SoH decline. These bundled interventions assess whether the resulting prediction changes align with the intended degradation semantics.
The optimistic intervention increases predicted SoH and achieves 69.0–90.5\% semantic consistency across the prediction horizons, while the pessimistic intervention decreases predicted SoH and achieves 100\% consistency. Both interventions generally shift SoH predictions in the expected direction based on domain knowledge, with stronger consistency observed for the pessimistic bundle. One possible explanation is that worsening degradation semantics are more compatible with the degradation trajectory already reflected in the observed time series, whereas imposing healthier rules may partially conflict with this temporal evidence.

\begin{table}[t]
\centering
\caption{Counterfactual analysis of different semantic interventions. Higher semantic consistency (\%) indicates that more predictions change in the expected direction after interventions. Positive (negative) $\bar{\Delta}_{H}$ values indicate an increase (decrease) in predicted SoH, with larger absolute values indicating stronger effects. $\bar{\Delta}_{H}$ values are in units of $10^{-3}$.}
\label{tab:counterfactual}
\footnotesize
\renewcommand{\arraystretch}{1.08}
\setlength{\tabcolsep}{2.5pt}
\begin{tabular}{lcccccccc}
\toprule
& \multicolumn{2}{c}{$H=30$}
& \multicolumn{2}{c}{$H=50$}
& \multicolumn{2}{c}{$H=70$}
& \multicolumn{2}{c}{$H=90$} \\
\cmidrule(lr){2-3}
\cmidrule(lr){4-5}
\cmidrule(lr){6-7}
\cmidrule(lr){8-9}
\textbf{Semantic Edit}
& \% & $\bar{\Delta}_{H}$
& \% & $\bar{\Delta}_{H}$
& \% & $\bar{\Delta}_{H}$
& \% & $\bar{\Delta}_{H}$ \\
\midrule
Optimistic Bundle
& 90.5 & $+$5.8 & 69.0 & $+$7.5
& 86.5 & $+$12.9 & 70.5 & $+$11.9 \\

Pessimistic Bundle
& \textbf{100.0} & $-$71.3
& \textbf{100.0} & $-$116.8
& \textbf{100.0} & $-$146.1
& \textbf{100.0} & $-$251.3 \\ \midrule

Slope $\to$ Mild
& 75.5 & $+$5.6 & 84.5 & $+$7.5
& 82.0 & $+$13.2 & 71.0 & $+$12.1 \\

Slope $\to$ Steep
& \textbf{100.0} & $-$68.9
& \textbf{100.0} & $-$129.7
& \textbf{100.0} & $-$142.3
& \textbf{100.0} & $-$255.1 \\

Trend $\to$ Stable
& 78.5 & $+$0.9 & 93.0 & $+$2.2
& 91.0 & $+$4.2 & 61.0 & $+$7.7 \\

Trend $\to$ Accel.
& 87.0 & $-$2.5 & 94.5 & $-$3.9
& 87.5 & $-$5.6 & 100.0 & $-$12.3 \\

Mech. $\to$ Stable
& 86.0 & $+$1.0 & 64.0 & $-$0.2
& 44.5 & $-$2.8 & 26.5 & $-$3.0 \\

Mech. $\to$ SEI
& 93.5 & $-$1.1 & 32.0 & $+$0.0
& 22.5 & $+$0.6 & 83.5 & $-$2.5 \\

Mech. $\to$ Li Plating
& 99.5 & $-$3.8 & 51.0 & $+$1.6
& 99.5 & $-$9.2 & 99.0 & $-$13.4 \\

Risk $\to$ Low
& 50.5 & $-$1.3 & 39.5 & $-$1.4
& 85.5 & $+$4.4 & 66.5 & $+$3.1 \\

Risk $\to$ High
& 19.0 & $+$2.6 & 34.5 & $-$0.1
& 58.5 & $+$0.2 & 37.0 & $-$0.3 \\
\bottomrule
\end{tabular}%
\end{table}

To distinguish the contributions of individual rules within the bundles, we next examine interventions that modify one semantic rule at a time. Among the individual interventions, Slope $\to$ Steep has the largest effect on the predicted SoH and achieves 100\% semantic consistency across all horizons. It also yields negative $\bar{\Delta}_H$ values, indicating lower predicted SoH when the slope is intervened toward steeper degradation. The mean SoH decrease grows from 6.9\% at $H=30$ to 25.5\% at $H=90$, indicating a stronger effect at longer prediction horizons as degradation effects accumulate over time. Trend $\to$ Accelerating also achieves high semantic consistency of 87.0–100.0\%, but produces smaller SoH decreases of 0.25–1.23\% only. These results indicate that, under the tested settings, the quantitative slope intervention has a stronger influence on the prediction than the categorical trend intervention. 
We also test mechanism interventions. For example, Mechanism $\to$ Li Plating produces negative $\bar{\Delta}_{H}$ values and high semantic consistency in most cases, as Li plating represents a detrimental degradation mechanism associated with accelerated capacity loss. For risk interventions, Risk $\to$ High achieves only 19.0–58.5\% semantic consistency and produces relatively small $\bar{\Delta}_{H}$ values, suggesting that its influence may depend more strongly on the battery state encoded in the temporal representation. Overall, the counterfactual analysis demonstrates that the learned semantic information meaningfully influences the SoH forecast and provides an interpretable characterization of the effects of different degradation semantics on the predictions.

\balance
\section{Conclusion}
\label{sec:conclusion}
This paper presents \textsc{Sera}, a framework for battery SoH forecasting that integrates structured text-based degradation semantics derived from battery domain knowledge with numerical battery time series. \textsc{Sera} employs a three-branch architecture that independently learns a temporal representation and rule- and LLM-based semantic representations, which are fused through adaptive gated aggregation for SoH forecasting. Compared with the temporal baseline, \textsc{Sera} reduces SoH forecasting errors by up to 37.3\% and improves generalization to unseen batteries. \textsc{Sera} also achieves consistent forecasting gains across different temporal encoders. Counterfactual semantic analysis provides insight into the influence of degradation semantics by examining the direction and magnitude of prediction changes under controlled interventions. The results show more consistent responses to semantic rules that directly characterize degradation progression. Overall, \textsc{Sera} improves SoH forecasting accuracy and interpretability by incorporating explicit degradation knowledge and semantics into temporal modelling effectively.

\balance
\begin{acks}
This research is supported by A*STAR under its MTC Individual Research Grants (IRG) (Award M23M6c0113), the Shanghai Sci-tech Co-research Program (Award 25HB2702600), and MTC Programmatic (Award M23L9b0052 and Award M24N6b0043).
\end{acks}

\bibliographystyle{ACM-Reference-Format}
\bibliography{SupportingFile/references}

\end{document}